\documentclass[10pt,twocolumn,letterpaper]{article}

\usepackage{cvpr}              

\usepackage{graphicx}
\usepackage{amsmath}
\usepackage{amssymb}
\usepackage{booktabs}

\usepackage[utf8]{inputenc} 
\usepackage[T1]{fontenc}    
\usepackage{url}            
\usepackage{booktabs}       
\usepackage{amsfonts}       
\usepackage{nicefrac}       
\usepackage{microtype}      
\usepackage{xcolor}         

\usepackage{graphicx}
\usepackage{verbatim}
\usepackage{multirow}
\usepackage{marvosym}
\usepackage{makecell}
\usepackage{bm}
\usepackage{subfloat}
\usepackage{wrapfig}
\usepackage{caption}
\usepackage{algorithm}
\usepackage{algorithmic}

\usepackage[accsupp]{axessibility}  

\newcommand{\R}{\ensuremath{\mathbb{R}}}

\def\x{\mathbf{x}}
\def\y{\mathbf{y}}

\usepackage[pagebackref,breaklinks,colorlinks]{hyperref}

\usepackage[capitalize]{cleveref}
\crefname{section}{Sec.}{Secs.}
\Crefname{section}{Section}{Sections}
\Crefname{table}{Table}{Tables}
\crefname{table}{Tab.}{Tabs.}

\def\cvprPaperID{4670} 
\def\confName{CVPR}
\def\confYear{2022}

\begin{document}

\title{Remember Intentions:
Retrospective-Memory-based Trajectory Prediction\vspace{-12pt}}

\author{
Chenxin Xu\textsuperscript{1\footnotemark[1]}, Weibo Mao\textsuperscript{1\footnotemark[1]}, Wenjun Zhang\textsuperscript{1}, Siheng Chen\textsuperscript{1,2\footnotemark[2]},
\\\textsuperscript{1}Shanghai Jiao Tong University,  \textsuperscript{2}Shanghai AI Laboratory
\\
{\tt\small \{xcxwakaka,kirino.mao,zhangwenjun,sihengc\}@sjtu.edu.cn}
}

\maketitle
\renewcommand{\thefootnote}{\fnsymbol{footnote}}
\footnotetext[1]{Equal contribution.\quad \quad \quad \footnotemark[2]Corresponding author.}\footnotetext{Code is available at: \url{https://github.com/MediaBrain-SJTU/MemoNet}}


\begin{abstract}
To realize trajectory prediction, most previous methods adopt the parameter-based approach, which encodes all the seen past-future instance pairs into model parameters. However, in this way, the model parameters come from all seen instances, which means a huge amount of irrelevant seen instances might also involve in predicting the current situation, disturbing the performance. To provide a more explicit link between the current situation and the seen instances, we imitate the mechanism of retrospective memory in neuropsychology and propose MemoNet, an instance-based approach that predicts the movement intentions of agents by looking for similar scenarios in the training data. In MemoNet, we design a pair of memory banks to explicitly store representative instances in the training set, acting as prefrontal cortex in the neural system, and a trainable memory addresser to adaptively search a current situation with similar instances in the memory bank, acting like basal ganglia. During prediction, MemoNet recalls previous memory by using the memory addresser to index related instances in the memory bank. We further propose a two-step trajectory prediction system, where the first step is to leverage MemoNet to predict the destination and the second step is to fulfill the whole trajectory according to the predicted destinations. Experiments show that the proposed MemoNet improves the FDE by 20.3\%/10.2\%/28.3\% from the previous best method on SDD/ETH-UCY/NBA datasets. Experiments also show that our MemoNet has the ability to trace back to specific instances during prediction, promoting more interpretability.

\end{abstract}
\vspace{-2mm}
\section{Introduction}
Trajectory prediction aims to predict the future movements for one or multiple interacting agents given the past trajectories. On the one hand, this task has broad practical applications to autonomous driving~\cite{levinson2011towards}, drones~\cite{floreano2015science}, surveillance systems~\cite{valera2005intelligent} and interactive robotics~\cite{kanda2002development}; on the other hand, this is a fundamental scientific question about linking the past to the future. The overall strategy is to summarize useful experiences from a large amount of seen past-future pairs and then leverage those experiences to predict possible future intentions for the current situation.

To obtain useful experiences, previous works consider a parameter-based approach, which uses training data to optimize model parameters. In this way, all the experiences are implicitly summarized and stored in a model as a whole during the optimization process. For example, \cite{mohamed2020social,huang2019stgat,yu2020spatio} use encoder-decoder architectures and~\cite{gupta2018social,hu2020collaborative} consider generator-discriminator architectures to regress future trajectory predictions. \cite{mangalam2020not,lee2017desire,ivanovic2019trajectron,salzmann2020trajectron++,yuan2021agentformer} use conditional variational autoencoders to sample multiple future trajectory embedding from latent distributions.~\cite{graber2020dynamic,li2020evolvegraph} rely on a bivariate Gaussian Mixture Model to output position distributions. However, the parameter-based approach is not optimal for two reasons. First, it lacks interpretability because all model parameters do not have clear semantic meaning in the physical world. This is critical in safety-sensitive applications, such as autonomous driving. Second, since the model parameters are trained from all seen instances, a huge amount of irrelevant seen past-future pairs might also involve in predicting the current situation, disturbing the performance.

\begin{figure}[t] 
\centering
\includegraphics[width=0.48\textwidth]{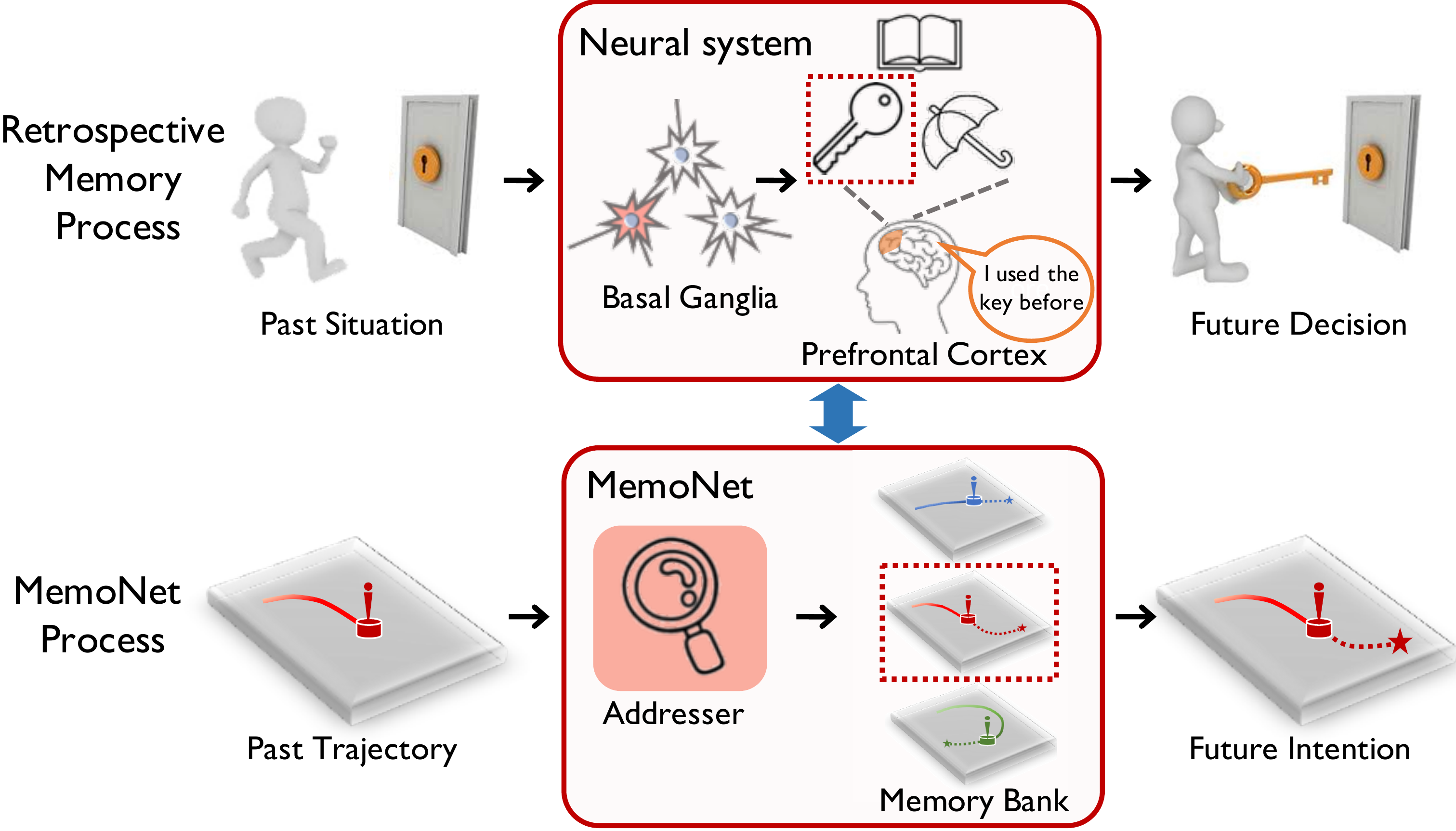}
\vspace{-7mm}
\caption{\small MemoNet mimics retrospective memory process. We use the memory bank to explicitly store representative instances, acting like prefrontal cortex; and the memory addresser to search similar memory instances with current situation, acting like basal ganglia.}
\label{fig:jewel}
\vspace{-15pt}
\end{figure}

To promote more interpretability and provide a more explicit link between the current situation and the seen instances, we propose \text{MemoNet} whose working mechanism is inspired by human's retrospective memory in neuropsychology \cite{goldman1995cellular,baddeley2013essentials}, the process that human learns intended future actions by recalling information learned before. The proposed MemoNet achieves the intention prediction by searching for similar instances stored during training. In MemoNet, we use a pair of past and intention memory banks to store the features of past-future instance pairs and a memory addresser to search relevant instances with the new prediction case in the memory bank. The memory bank simulates the prefrontal cortex in the neural system, which records the human reaction when performing a task. The memory addresser simulates the basal ganglia in the neural system, which activates the related memory records in the prefrontal cortex. Fig.\ref{fig:jewel} shows an analogy between the retrospective memory process and our MemoNet process.

The proposed MemoNet includes four key designs. First, we propose a joint-reconstruction-based feature-learning architecture to initialize the pair of past and intention memory banks. The architecture contains two encoders and follows a joint-reconstruction structure to obtain compatible past trajectory and future intention features. Second, we propose a memory filter algorithm to erase the redundant memory instances in the memory banks. The filter algorithm is training-free and invariant to the permutation of training samples, providing high efficiency and robustness for the memory banks. Third, we propose a trainable memory addresser to search similar memory instances. The addresser contains a learnable attention network to compute similarity scores. To train such an addresser, we propose a pseudo-label generation to guide the addresser to correctly search most similar memory instances. Fourth, we propose an intention clustering to produce diverse intention predictions. Through the clustering algorithm, intentions with low-frequency occurrences are captured to promote the prediction diversity and intentions with high-frequency occurrences are merged to improve the prediction robustness.

We build a two-step trajectory prediction system, where the first step is to leverage MemoNet to predict the intentions and the second step is to fulfill the whole trajectory according to the predicted intentions. Note that MemoNet only predicts the destination to represent the intention because the destination carries most of the modality information in a trajectory. This two-step prediction disassembles a complex problem into two relatively simple problems, promoting a more accurate prediction. To evaluate the effectiveness of our method, we conduct experiments on three datasets: Stanford Drones (SDD), ETH-UCY and NBA. The quantitative result shows we outperform the previous state-of-the-art method 20.3\%/10.2\%/28.3\% on FDE representing we achieve an accurate intention prediction with the MemoNet. The qualitative results also reflect that our MemoNet has the ability to trace back to specific memorized samples during the prediction, promoting more interpretability.

The main contributions of this paper are:

$\bullet$ We propose MemoNet, a novel instance-based framework to achieve future intention prediction. The working mechanism of MemoNet is based on a more explicit link between the current situation and seen instances, imitating retrospective memory studied in neuropsychology.


$\bullet$ We propose four novel designs in MemoNet, including 1) reconstruction-based feature-learning architecture, which initializes the memory banks, 2) memory filtering, which reduces the redundancy in memory banks, 3) memory addresser, which searches similar memory instances with the incoming prediction case in memory banks, and 4) intention clustering, which promotes prediction diversity.


$\bullet$ We conduct experiments to evaluate our method on several real-world datasets. Our approach achieves the state-of-the-art on well-established pedestrian trajectory prediction datasets by reducing the FDE 20.3\%/10.2\%/28.3\% on SDD/ETH-UCY/NBA datasets. Our approach also equips with the ability to trace back to specific memorized instances during the prediction, promoting more interpretability.

\section{Related Work}
\textbf{Trajectory prediction.} 
Early work on trajectory prediction adopts a deterministic approach using models such as social forces \cite{helbing1995social,mehran2009abnormal}, Markov process \cite{kitani2012activity,wang2007gaussian}, and RNNs \cite{alahi2016social,morton2016analysis,vemula2018social}. 
Recently, researchers begin to propose frameworks to predict multi-model trajectories, which can be mainly categorized into two types: regression, generation. Regression frameworks mainly utilize encode-decode structures \cite{mohamed2020social,huang2019stgat,yu2020spatio,li2021online,gao2020vectornet}, or reinforcement learning-based structure\cite{0001YMMTC21},  or generator-discriminator structures \cite{gupta2018social,hu2020collaborative} with adding noise \cite{mohamed2020social,huang2019stgat,yu2020spatio,gupta2018social,hu2020collaborative}, using random initialization \cite{liang2019peeking}, or using multi-head output \cite{liang2020learning,tang2021collaborative} to regress multiple future trajectories. Generation frameworks estimate the distribution of future trajectory or its embedding with deep generative models \cite{kingma2013auto}.  \cite{graber2020dynamic,li2020evolvegraph} utilize a Gaussian mixture distribution to model the future trajectory distribution and the model estimates its mean and covariance. The mainly used framework is conditional variational autoencoders~\cite{mangalam2020not,lee2017desire,ivanovic2019trajectron,salzmann2020trajectron++,yuan2021agentformer}, which achieve the prediction by estimating the parameters of an intermediate distribution and sampling future trajectory features from such a distribution.

Both the regression and generation frameworks are parameter-based, utilizing training data to optimize model parameters. In such frameworks, learned experience is a hidden representation stored implicitly in model parameters as a whole, lacking the ability to address an individual instance of experience. In this work, we propose a new instance-based framework based on retrospective memory which memorizes various past trajectories and corresponding intentions. In predicting, the framework recalls similar previous memory instances for guiding future prediction. Compared with previous methods, our method provides a more explicit link between the current prediction and seen data, which promotes more interpretability and higher performance.

\textbf{Memory Networks.}
The first proposed memory network is called Neural
Turing Machines (NTM) \cite{graves2014neural} which is analogous to a Von Neumann architecture consisting of a neural network controller and a memory bank. The NTM architecture is extended in meta-learning \cite{santoro2016meta} which implements a Least Recently Used memory access strategy to make predictions using few samples. \cite{graves2016hybrid} proposes a differentiable neural computer that can read from and write to an external memory matrix. Memory network is also proved its effectiveness on question-answering tasks \cite{weston2014memory} where the model stores the question-answering pair into a long-term memory as a knowledge base and outputs a textual response. \cite{sukhbaatar2015end} proposes an end-to-end memory network for question-answering with a recurrent attention model in which the recurrence reads from a large external memory. \cite{kumar2016ask,ma2018visual} apply memory networks further into visual question-answering tasks \cite{antol2015vqa}. \cite{MaSLTC21} applies a generative memory for continual trajectory prediction. 

A close related work with ours is \cite{marchetti2020mantra}, which leverages the memory mechanism to achieve single-agent trajectory prediction. However, the differences include four aspects: i) the previous work only considers single-agent trajectory prediction; while the proposed MemoNet is able to handle multi-agent trajectory prediction with social influence; ii) the memory bank in the previous work stores the entire trajectories; while MemoNet focuses on intention, which is more efficient in memorizing possible movement patterns; iii) the previous work uses fixed cosine similarity to search related memories; while MemoNet uses a trainable addresser to learn a similarity metric, leading to better memory searching; and iv) the previous work is hard to both ensure diversity and preserve precision while MemoNet adopts intention clustering to promote multi-modality prediction with robustness. Overall, the proposed MemoNet outperforms~\cite{marchetti2020mantra} by 28.7\%/46.2\% in FDE on SDD/ETH-UCY datasets.

\begin{figure*}[t] 
\centering
\includegraphics[width=0.98\textwidth]{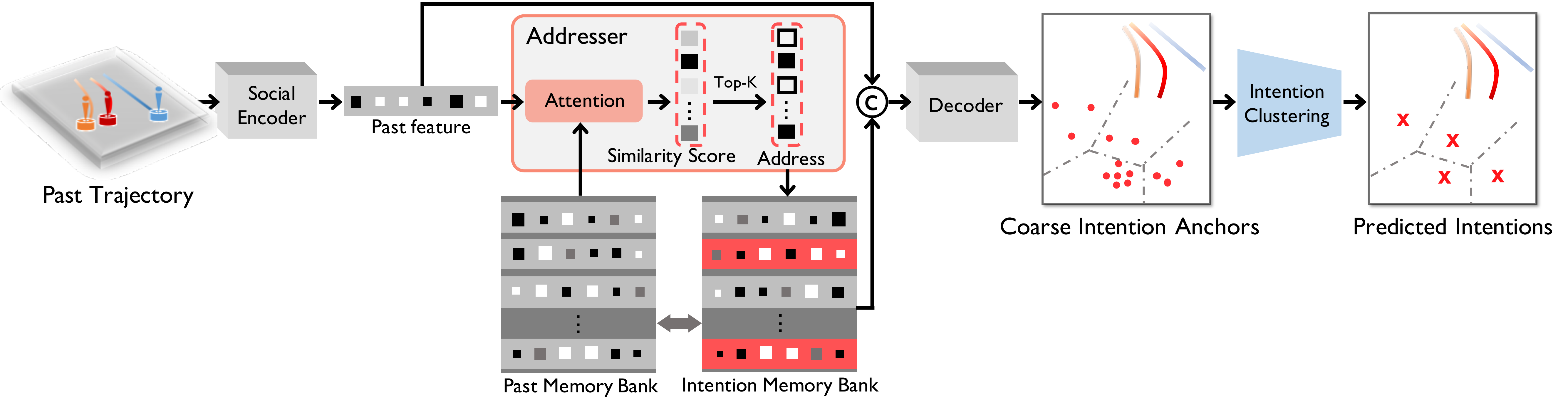}
\vspace{-3mm}
\caption{\small Inference phase of MemoNet. The red agent is to-be-predicted and the blue/orange agents are neighbours. According to the past feature obtained by the social encoder, we address related memory instances in the past memory bank through an attention network, producing similarity scores. The intention memory bank outputs future intention features for decoding coarse intention anchors according to the memory addresses with top similarity scores. At last, we utilize a clustering algorithm to obtain diverse and robust intention predictions.}
\label{fig:memonet}
\vspace{-15pt}
\end{figure*}

\section{Problem Formulation}
Trajectory prediction is to predict an agent's future trajectory from its past trajectory and neighboring agents' past trajectories. 
Mathematically, for a to-be-predicted agent, let $\x^t \in \R^2$ be its spatial coordinate at timestamp $t$ and $\mathbf{X} = [\x^{-T_{\rm p}+1}, \x^{-T_{\rm p}+2},\cdots, \x^{0}] \in \R^{T_{\rm p} \times 2}$ be its past trajectory over $T_{\rm p}$ timestamps. Let $\mathcal{N}$ be the neighbouring agent set and $\mathbb{X}_{\mathcal{N}} = [\mathbf{X}_{\mathcal{N}_1} ,\mathbf{X}_{\mathcal{N}_2},\cdots,\mathbf{X}_{\mathcal{N}_N}] \in \R^{N \times T_p \times 2}$ be the past trajectories of neighbours, where $\mathbf{X}_{\mathcal{N}_\ell}\in \R^{T_{\rm p} \times 2}$ is the trajectory of the $\ell$th neighbour. The future trajectory of the to-be-predicted agent is $\mathbf{Y} = [\y^{1}, \y^{2},\cdots, \y^{T_{\rm f}}] \in \R^{T_{\rm f} \times 2}$ where $\y^t \in \R^2$ is the spatial coordinate of at future timestamp $t$. The overall goal is to train a prediction model $g(\cdot)$, so that the predicted future trajectory $\widehat{\mathbf{Y}} = g(\mathbf{X}, \mathbb{X}_{\mathcal{N}})$ is as close to the ground-truth $\mathbf{Y}$ as possible. 

To reach this goal, we consider a two-step strategy, where we first predict the agent's intention and then fulfill the complete trajectory based on the predicted intention. The intuition behind is to disassemble a complex problem into two relatively simple problems, promoting a more accurate prediction. Here we represent the agent's intention by its destination as the destination could reflect most of the movement patterns. Mathematically, we target to learn an intention prediction model $g_\mathrm{int}(\cdot)$ that predicts a intention $\widehat{\mathbf{y}}^{T_f} = g_\mathrm{int}(\mathbf{X},\mathbb{X}_\mathcal{N})$.  We next target to train the trajectory fulfilling  model $g_\mathrm{full}(\cdot)$ based on the predicted intention $\widehat{\mathbf{Y}} = g_\mathrm{full}(\mathbf{X},\mathbb{X}_\mathcal{N},\widehat{\mathbf{y}}^{T_f})$.  In this spirit, we propose MemoNet for intention prediction; see Sec.\ref{Sec:MemoNet}; we then build the overall prediction model based on MemoNet; see Sec.\ref{Sec:system}. 

\section{MemoNet: Intention Prediction}
\label{Sec:MemoNet}
MemoNet exploits retrospective memory from similar scenarios of previous experience to obtain the possible multimodal future movement intentions. The core of MemoNet is to store representative instances in the memory bank and then use a memory addresser to search relevant seen instances with the current situation in the memory bank. Sec.~\ref{sec:Memory learning} proposes the memory bank and Sec.~\ref{sec:addresser} proposes the memory addresser. To enable diverse intention prediction, we propose intention clustering in Sec.~\ref{sec:Destination clustering}. Finally, we summarize the inference process of MemoNet in Sec.~\ref{sec:inference}.

\subsection{Memory bank}
\label{sec:Memory learning}
\vspace{-1mm}
\textbf{Memory bank initialization.} We consider a pair of correlated memory banks: a past memory bank and an intention memory bank. The past memory bank stores a set of past trajectory features and the intention memory bank stores a set of corresponding future intention features. They together associate the past with the future. Mathematically, let $\mathcal{M}_\mathrm{past}=\{\mathbf{k}_i | i=1,2,\cdots,M\}$ be the past memory bank, where $\mathbf{k}_{i}$ is the instance at the $i$th memory address, recording the features extracted from the past trajectory with social influence in the $i$th training sample. Correspondingly, $\mathcal{M}_\mathrm{int}=\{\mathbf{v}_i | i=1,2,\cdots,M\}$ be the intention memory bank, where $\mathbf{v}_{i}$ is the instance at the $i$th address, recording the features extracted from the future intention (destination) in the $i$th training sample. Both the past and the intention memory banks share the same size $M$. 

\begin{figure*}[!t]
\centering
\subfloat[\small Feature learning architecture via joint reconstruction.]{
\begin{minipage}[t]{0.44\linewidth}
\centering
\includegraphics[height=0.4\textwidth]{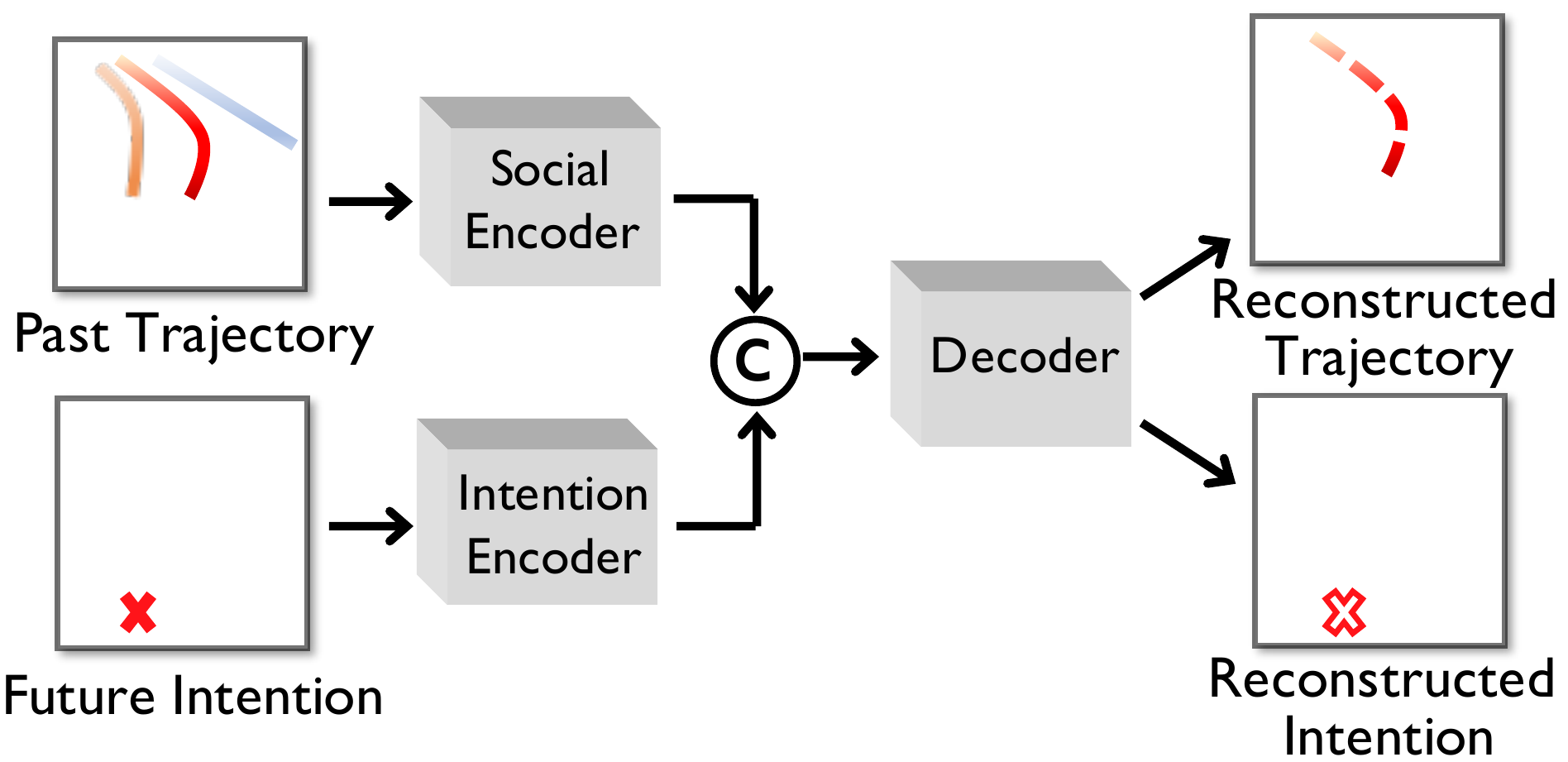}
\end{minipage}%
}
\subfloat[\small Initialization process via enumerating past-intention samples.]{
\begin{minipage}[t]{0.47\linewidth}
\centering
\includegraphics[height=0.4\textwidth]{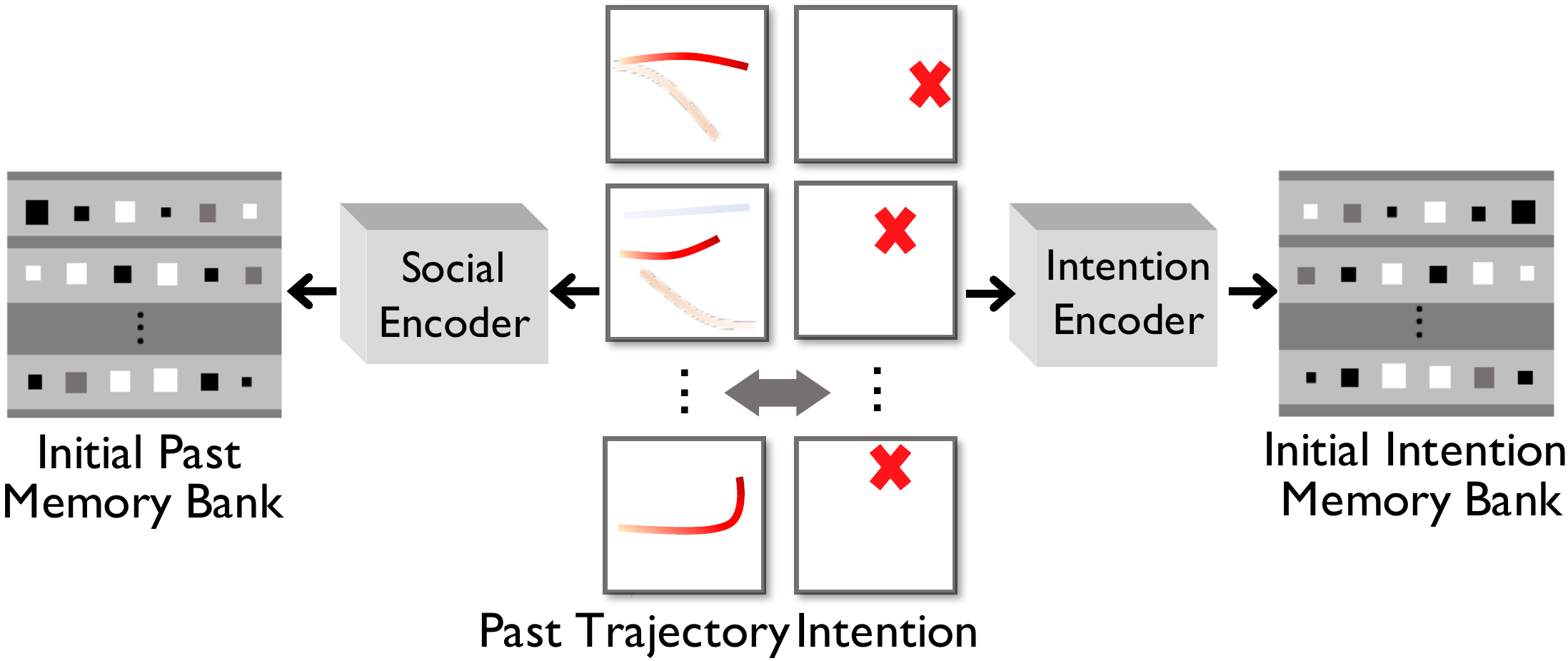}
\end{minipage}%
}%
\centering
\vspace{-12pt}
\caption{\small Memory bank initialization. We train a feature learning architecture by a joint-reconstruction process and initialize the memory bank by enumerating all past-intention samples using two encoders.}
\label{fig:reconstruction}
\vspace{-15pt}
\end{figure*}

To obtain the features in the memory bank pair $\mathbf{k}_{i},\mathbf{v}_{i}$, we propose a joint-reconstruction-based feature learning architecture; see Fig \ref{fig:reconstruction}(a). 
The social encoder extracts the past feature with social influence of the past trajectory. The intention encoder extract the intention feature from the future intention (destination). The decoder receives the concatenated past-and-intention features and reconstructs the past trajectory and the future intention jointly. 
Mathematically, let $\mathcal{E}_\mathrm{social}(\cdot)$ and $\mathcal{E}_\mathrm{int}(\cdot)$ be the social encoder and intention encoder, $\mathcal{D}(\cdot)$ be the decoder, given an agent's trajectory $\mathbf{X}$, its neighbouring agents' trajectories $\mathbb{X}_{\mathcal{N}}$, and its future intention $\mathbf{y}^{T_f}$, the joint-reconstruction process is:
\begin{equation*}
   \setlength{\abovedisplayskip}{3pt}
   \setlength{\belowdisplayskip}{3pt}
    \mathbf{k} = \mathcal{E}_\mathrm{social}(\mathbf{X}, \mathbb{X}_{\mathcal{N}}),\; \mathbf{v} = \mathcal{E}_\mathrm{int}(\mathbf{y}^{T_f}),\;  \widehat{\mathbf{X}},\widehat{\mathbf{y}}^{T_f} = \mathcal{D}([\mathbf{k};\mathbf{v}]),
\end{equation*}
where $[\cdot;\cdot]$ represents the concatenate operation and $\widehat{\mathbf{X}},\widehat{\mathbf{y}}^{T_f}$ denote the reconstructed past trajectory and future intention.

To optimize the feature learning architecture, we use a joint-reconstruction loss function:
\begin{equation*}
   \setlength{\abovedisplayskip}{3pt}
   \setlength{\belowdisplayskip}{3pt}
    \mathcal{L}_{\mathrm{rec}} = \|\widehat{\mathbf{X}}-\mathbf{X}\|_{2}^{2} + \alpha \left\|\widehat{\mathbf{y}}^{T_f}-\mathbf{y}^{T_f}\right\|_2^{2},
    \label{equ:loss_reconstruction}
\end{equation*}
where $\alpha$ is a weight hyperparameter. Through the proposed feature learning architecture, we obtain respective feature of the past and the intention. Their features are compatible because of the joint-reconstruction process.

Once we finish the feature learning architecture, we fix the past and the intention encoders and enumerate over all the past-intention samples in training data to initialize the past memory bank $\mathcal{M}_\mathrm{past}^{(0)}$ and the intention memory bank $\mathcal{M}_\mathrm{int}^{(0)}$. Specifically, for the $i$th past/intention sample, we use the social encoder/intention encoder to get the past feature $\mathbf{k}_i$/intention feature $\mathbf{v}_i$  storing at the $i$th address of the past/intention memory bank; see Fig.\ref{fig:reconstruction}(b).

\textbf{Memory bank filtering.}
When we write all the past and intention features into the memory bank pair, many instances could be redundant, which wastes the storage. We thus propose a filtering algorithm to erase redundant memory instances and preserve representative memory instances.

For features $\mathbf{k}_i$,$\mathbf{v}_i$ at $i$th address in initial memory bank pair $\mathcal{M}_\mathrm{past}^{(0)}$ and $\mathcal{M}_\mathrm{int}^{(0)}$, we use its corresponding starting position and intention $\mathbf{x}_{i}^{-T_p+1}, \mathbf{y}_i^{T_f}$ to filter similar memory instances. For the $i$th and the $j$th addresses, if their memory instances have close past starting positions and future intentions, this pair of addresses is redundant and one should be removed. Mathematically, for memory instances in the $i$th address with its starting position $\mathbf{x}_{i}^{-T_p+1}$ and intention $\mathbf{y}_i^{T_f}$ and the $j$th address with its starting position $\mathbf{x}_{j}^{-T_p+1}$ and intention $\mathbf{y}_j^{T_f}$, they are redundant when:
\begin{equation}
   \setlength{\abovedisplayskip}{3pt}
   \setlength{\belowdisplayskip}{3pt}
    \|\mathbf{x}_{i}^{-T_p+1}-\mathbf{x}_{j}^{-T_p+1}\|_2 \leq \theta_{\rm past},\; \|\mathbf{y}_i^{T_f}-\mathbf{y}_j^{T_f}\|_2 \leq \theta_{\rm int},
    \label{equ:filter}
\end{equation}
where $\theta_{\rm past}$ and $\theta_{\rm int}$ are two thresholds for tuning. We use this rule to filter the past and the intention memory bank; see Algorithm \ref{alg1}. Briefly, $\theta_{\rm past}/\theta_{\rm int}$ will control the memory size of the final past/intention memory banks.

Compared to previous method that uses a controller to reduce redundancy~\cite{marchetti2020mantra}, our filtering has two advantages. First, our memory bank is invariant to the permutation of training samples; while in the previous method, various orderings of training samples would cause unstable memory banks. Second, our memory filter is training-free, which is more efficient; while the previous method needs to train the controller for multiple epochs. 

\textbf{Relations to previous methods.}
The proposed memory bank is similar to dictionary learning as both aim to infer a few representatives from input data to approximate incoming data, but differences include: i) a dictionary usually requires a fixed and predefine size; while the size of a memory bank is flexible and adaptive to the complexity of input data; ii) to make a prediction, the dictionary usually combines several atoms by weighted averaging; while the memory bank directly searches a single memory instance that allows an explicit link between the inference data and the training data.

\setlength{\textfloatsep}{4pt}
\begin{algorithm}[t]
\algsetup{linenosize=\small} \small
\caption{\small Memory bank filtering} 
\label{alg1}
\begin{algorithmic}[1]
\vspace{-1mm}
\REQUIRE ~~ Initial memory banks $\mathcal{M}_\mathrm{past}^{(0)}$, $\mathcal{M}_\mathrm{int}^{(0)}$\\
\ENSURE Filtered memory banks $\mathcal{M}_\mathrm{past}$,$\mathcal{M}_\mathrm{int}$
\STATE Initialize $\mathcal{M}_\mathrm{past}=\varnothing$,$\mathcal{M}_\mathrm{int}=\varnothing$ 
\WHILE{$\mathcal{M}_\mathrm{past}^{(0)}\ne \varnothing$ and $\mathcal{M}_\mathrm{int}^{(0)}\ne \varnothing$}
\STATE Randomly pick address $i$ in $\mathcal{M}_\mathrm{past}$, $\mathcal{M}_\mathrm{int}$
\STATE {\textbf{for} all address $j$ in current $\mathcal{M}_\mathrm{past}$,$\mathcal{M}_\mathrm{past}$}
\IF {Eq.(\ref{equ:filter}) not satisfied for all addresses $j$ }
\STATE Add $\mathbf{k}_i$,$\mathbf{v}_i$ into $\mathcal{M}_\mathrm{past}$,$\mathcal{M}_\mathrm{int}$
\STATE Delete $\mathbf{k}_i$,$\mathbf{v}_i$ from $\mathcal{M}_\mathrm{past}^{(0)}$,$\mathcal{M}_\mathrm{int}^{(0)}$
\ENDIF
\ENDWHILE
\RETURN $\mathcal{M}_\mathrm{past}$,$\mathcal{M}_\mathrm{int}$
\end{algorithmic}
\vspace{-1mm}
\end{algorithm}

\subsection{Memory addresser}
\label{sec:addresser}
\vspace{-1mm}
The functionality of a memory addresser is to search the addresses of similar past memory instances in the memory bank for an input past trajectory feature. The key is to find an appropriate similarity metric. The previous memory addressing mechanisms leverage the cosine distance between two features as the similarity metric~\cite{graves2014neural,marchetti2020mantra}. However, any pre-defined function, including the cosine distance, might not be capable of fully reflecting the similarity between two feature vectors. To solve this issue, we propose a trainable addresser, which contains a shallow attention network to learn a similarity metric.
Mathematically, given the input past feature $\mathbf{q}$ and the past memory bank $\mathcal{M}_\mathrm{past}=\{\mathbf{k}_i|i=1,2,\cdots,|\mathcal{M}|\}$, we calculate the similarity scores across all the memory instances, which is formulated as:
\begin{equation*}
   \setlength{\abovedisplayskip}{4pt}
   \setlength{\belowdisplayskip}{5pt}
    s_i = \mathcal{F}_\mathrm{ATT}(\mathbf{q},\mathbf{k}_i) = \frac{\mathcal{F}_{\mathrm{q}}(\mathbf{q}) \mathcal{F}^\mathrm{T}_{\mathrm{k}}(\mathbf{k}_i)}{\|\mathcal{F}_{\mathrm{q}}(\mathbf{q}) \|_2 \| \mathcal{F}_{\mathrm{k}}(\mathbf{k}_i) \|_2}, i=1,2,\cdots, M,
\end{equation*}
where $\mathcal{F}_q(\cdot)$ and $\mathcal{F}_k(\cdot)$ are two individual MLPs that transform features to a space for more appropriate distance measuring, $s_i$ is the similarity score between the input feature and the $i$th memory instance. We then select the largest similarity scores and return their memory addresses.

\begin{figure}[t] 
\centering
\includegraphics[width=0.45\textwidth]{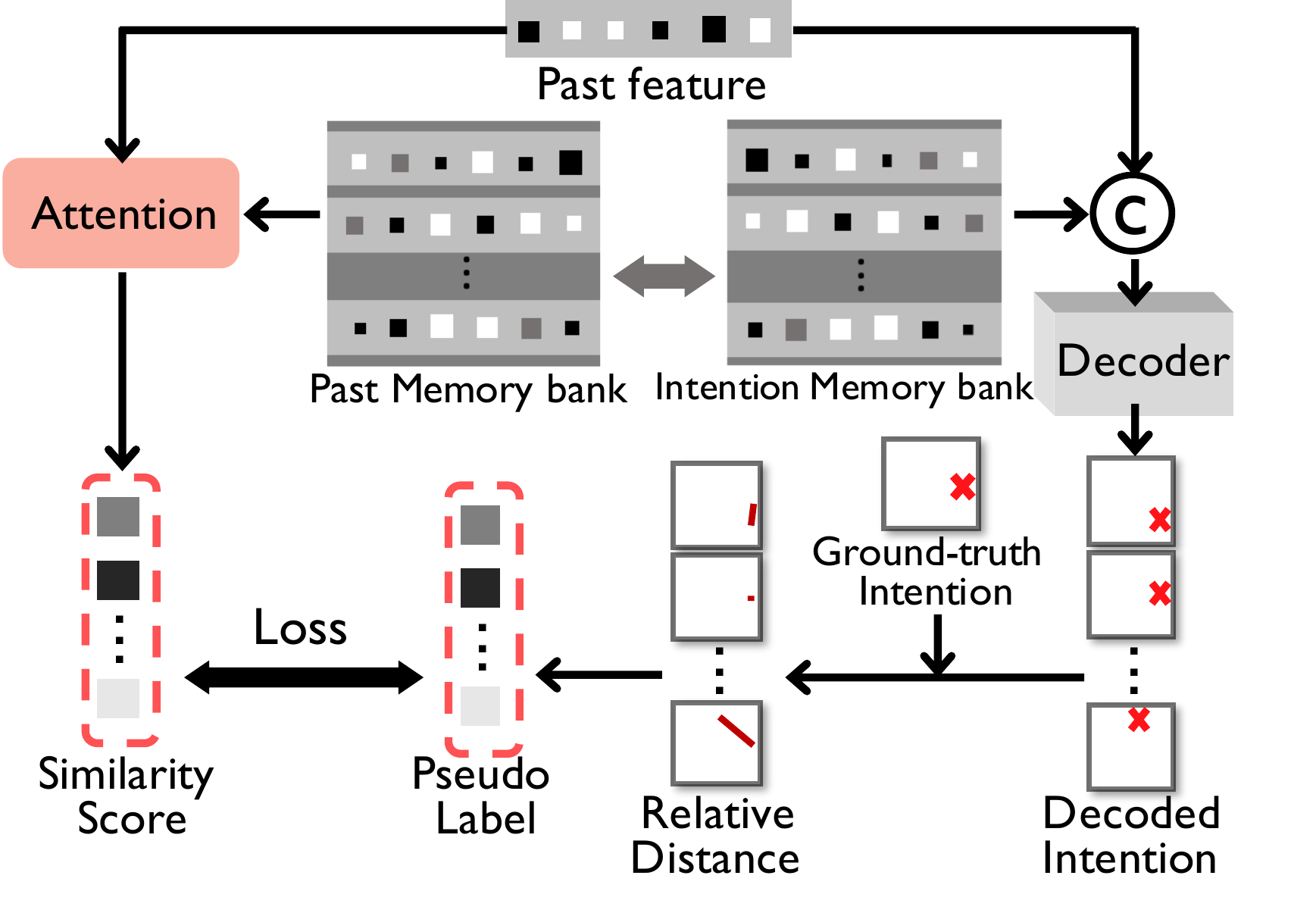}
\vspace{-3mm}
\caption{\small The addresser learning process. To train the attention network, we generate a pseudo label based on the relative distance between the decoded intentions and the ground-truth intentions.}
\label{fig:address}
\vspace{-0pt}
\end{figure}

To train such an addresser, we need to determine the "ground-truth" similarity score. 
Intuitively, the similarity score measured in the feature space should reflect the prediction error in the physical space. We thus consider a pseudo label which is related to the relative distance between the ground-truth intention of the input and the predicted intentions. Mathematically, let $\mathbf{y}^{T_f}$ be the ground-truth intention of the input trajectory and $\widehat{\mathbf{y}}_i^{T_f} = \mathcal{D}([\mathbf{k}_i;\mathbf{v}_i])$ be the predicted intention of the $i$th memory instance produced by the aforementioned intention decoder $\mathcal{D}(\cdot)$. The pseudo label the $i$th memory instance is defined as $\mathrm{max}(0,\frac{d_\mathrm{T}-d_\mathrm{i}}{d_\mathrm{T}}) \in [0,1]$, where $d_i = \|\mathbf{y}^{T_f}-\widehat{\mathbf{y}}_i^{T_f}\|_2$ is the relative distance between two intentions and $d_T$ is a distance threshold. 
Based on this pseudo label, we train the addresser with the following loss:
\begin{equation*}
   \setlength{\abovedisplayskip}{1pt}
   \setlength{\belowdisplayskip}{1pt}
    \mathcal{L}_\mathrm{Addr} = \sum_{i=1}^{M} (s_i-\mathrm{max}(0,\frac{d_\mathrm{T}-d_\mathrm{i}}{d_{T}}))^2;
\end{equation*}
see the training process of a memory addresser in Figure~\ref{fig:address}.


\subsection{Intention diversity}
\label{sec:Destination clustering}
\begin{figure}[!t]
\centering
\subfloat[\small No intention clustering]{
\begin{minipage}[t]{0.45\linewidth}
\flushleft
\includegraphics[width=0.95\textwidth]{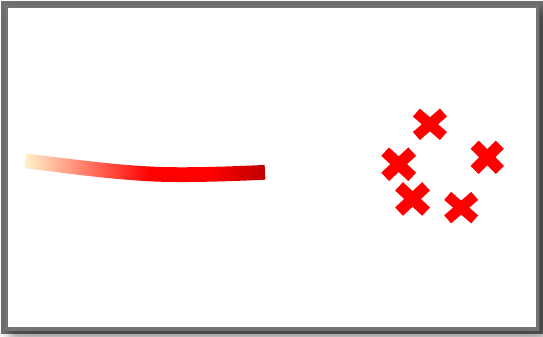}
\end{minipage}%
}%
\subfloat[\small With intention clustering]{
\begin{minipage}[t]{0.45\linewidth}
\flushright
\includegraphics[width=0.945\textwidth]{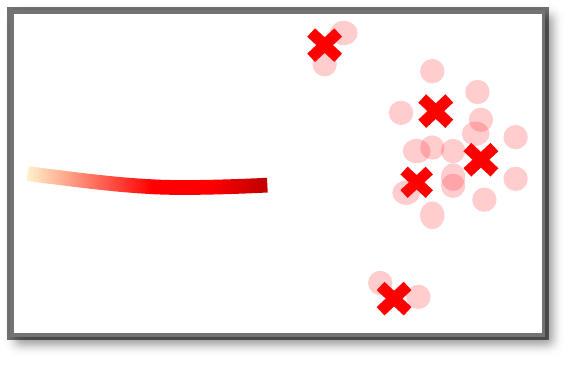}
\end{minipage}%
}%
\centering
\vspace{-13pt}
\caption{\small Examples of intention prediction. With intention clustering, MemoNet produces a more diverse prediction.}
\label{fig:clustering}
\vspace{-1pt}
\end{figure}

\vspace{-1mm}
Fig.~\ref{fig:clustering}(a) illustrates a scenario that the top few searched memory instances might fall into the same modality and cannot provide sufficient diversity. The reason is that the memory bank might recall numerous seen instances like the agent will move straight in various ways, but miss other movement modalities, such as sharp left turn or right turn. Note that although simply using memory bank filtering with a large $\theta_\mathrm{past}$/$\theta_\mathrm{int}$ could promote diversity, it would remove too many memory instances and make it harder to search relevant memory instances, deteriorating the performance. To achieve a diverse prediction preserving precision, we propose an intention clustering method.

Suppose that we need to predict $K$ possible trajectories. Here we first find $L$ ($L \gg K$) memory instances based on the $L$ largest similarity scores and then decode them into $L$ intentions, which are called coarse intention anchors. We then use K-means clustering method to produce $K$ possible intentions from the $L$ coarse intention anchors. On one hand, since $L \gg K$, the coarse intention anchors are more likely to capture more agents' movement patterns and the clustering operation is capable to preserves these patterns to produce a more diverse prediction. 
On the other hand, intention clustering preserves the enrichment of the memory bank and considers multiple memory instances to cluster a predicted intention, leading to a more precise and confident intention prediction, see the example in Fig.\ref{fig:clustering}(b).

\subsection{Inference Phase}
\label{sec:inference}
\vspace{-1mm}
During inference, MemoNet involves four steps to obtain possible future intentions: past trajectory encoding, memory addressing, intention decoding and intention clustering; see Fig.\ref{fig:memonet}.  First, we input the past trajectory and its neighbouring past trajectories to the social encoder $\mathcal{E}_\mathrm{social}(\cdot)$ to obtain the past trajectory feature. Second, we search $L$ most related memory instances in the past memory bank through the proposed memory addresser and return their addresses. According to the memory addresses, the intention memory bank outputs $L$ corresponding future intention features. Third, the decoder $\mathcal{D}(\cdot)$ decodes each of $L$ intention features into $L$ intention anchors. Fourth, we use the proposed clustering algorithm to refine $L$ intention anchors to $K$ final intentions.

\section{Trajectory prediction system}
\label{Sec:system}
\subsection{Trajectory fulfilling}
\label{sec:trajectory prediction}
\begin{figure}[t] 
\centering
\includegraphics[width=0.47\textwidth]{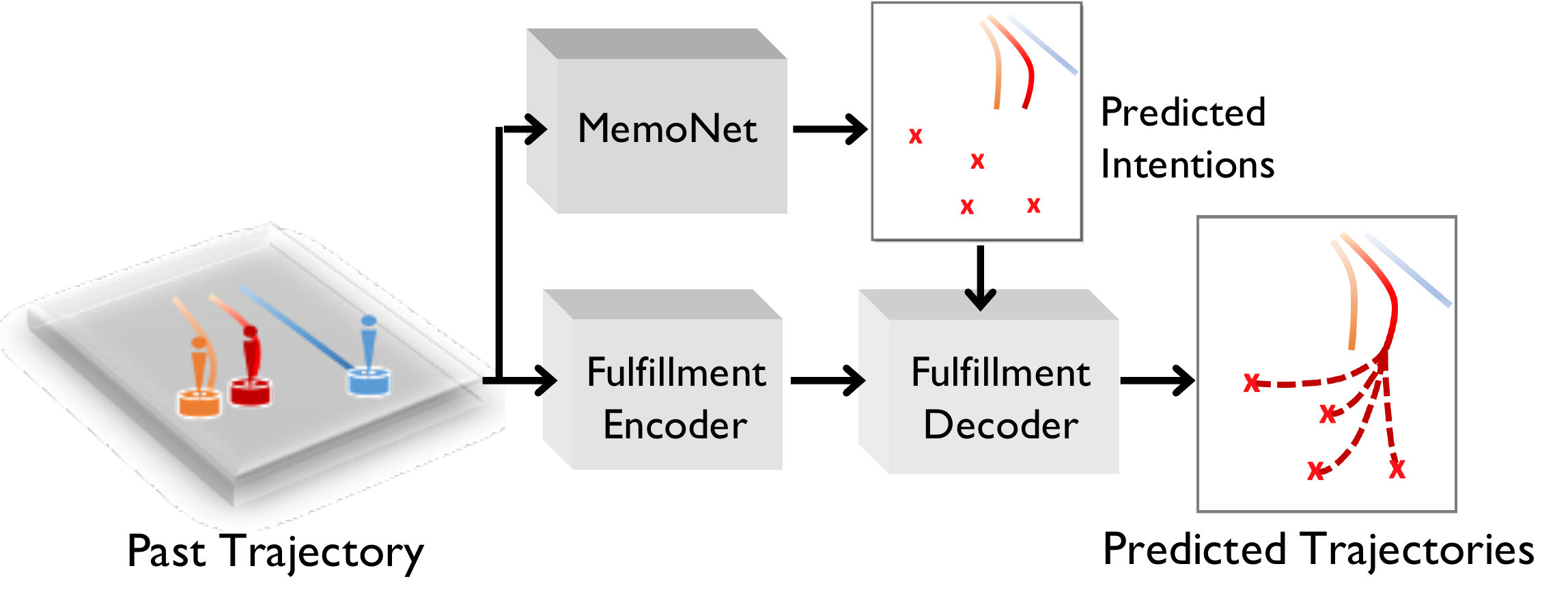}
\vspace{-3mm}
\caption{\small The inference of trajectory prediction system with MemoNet. The red color represents the to-be-predict agent and the blue/red color represent neighbours. We fulfill the whole trajectory conditioned on the prediction intentions from MemoNet.}
\label{fig:system}
\vspace{-1pt}
\end{figure}

\vspace{-1mm}
After obtaining an agent's trajectory intentions (destinations), we fulfill the whole trajectory through an encoding-decoding process conditioned on predicted intentions; see Fig.\ref{fig:system}. Mathematically, given a predicted intention $\widehat{\mathbf{y}}^{T_f}$ of the agent's past trajectory $\mathbf{X}$ with its neighbours' past trajectories $\mathbb{X}_\mathcal{N}$, the trajectory fulfilling process is:
\begin{equation*}
  \setlength{\abovedisplayskip}{4pt}
  \setlength{\belowdisplayskip}{4pt}
 \begin{aligned}
    \mathbf{h}_\mathrm{x} = {\mathcal{E}}_\mathrm{full}(\mathbf{X}&,~\mathbb{X}_\mathcal{N}),~ \mathbf{h}^{\prime}_x = [\mathbf{h}_\mathrm{x};~\mathcal{F}_\mathrm{d}(\widehat{\mathbf{y}}^{T_f})],~ \\ \widehat{\mathbf{Y}},~
    &\widehat{\mathbf{X}}_\mathrm{full} = {\mathcal{D}}_\mathrm{full}(\mathbf{h}^{\prime}_x),
\end{aligned}
\end{equation*}
where ${\mathcal{E}}_\mathrm{full}(\cdot)$ and ${\mathcal{D}}_\mathrm{full}(\cdot)$ represent the trajectory fulfillment encoder and decoder, which share a same structure with $\mathcal{E}_\mathrm{social}(\cdot)$ and $\mathcal{D}(\cdot)$, respectively. We concatenate the trajectory feature $\mathbf{h}_\mathrm{x}$ with intention feature encoded by a MLP function $\mathcal{F}_\mathrm{d}(\cdot)$ for whole trajectory $\widehat{\mathbf{Y}}$ decoding. To keep most past information, the fulfillment decoder also aim to reconstruct the past trajectory $\widehat{\mathbf{X}}_\mathrm{full}$. To train the fulfillment encoder and decoder, we use the $\ell_2$ loss:
\begin{equation*}
   \setlength{\abovedisplayskip}{4pt}
   \setlength{\belowdisplayskip}{4pt}
    \mathcal{L}_\mathrm{traj} = \|\widehat{\mathbf{X}}_\mathrm{full}-\mathbf{X}\|_2^2+\beta \|\widehat{\mathbf{Y}}-\mathbf{Y}\|_2^2,
\end{equation*}
where $\beta$ is a weight hyperparamter.

\subsection{Overall training pipeline}
\label{sec:pipeline}
\vspace{-1mm}
To train the overall system with MemoNet, we design the following training pipeline:

1. Train two encoders $\mathcal{E}_\mathrm{social}(\cdot)$, $\mathcal{E}_\mathrm{int}(\cdot)$ and the decoder $\mathcal{D}(\cdot)$ using the feature learning architecture with the joint-reconstruction loss $\mathcal{L}_\mathrm{rec}$. 

2. Freeze two encoders $\mathcal{E}_\mathrm{social}(\cdot)$, $\mathcal{E}_\mathrm{int}(\cdot)$. Create the pair of initial past and intention memory banks $\mathcal{M}^{(0)}_{\rm past}$ and $\mathcal{M}^{(0)}_{\rm int}$ by using $\mathcal{E}_\mathrm{social}(\cdot)$, $\mathcal{E}_\mathrm{int}(\cdot)$. Apply memory filtering to obtain the final past and intention memory banks $\mathcal{M}_{\rm past}$ and $\mathcal{M}_{\rm int}$.

3. Freeze the memory banks $\mathcal{M}_{\rm past}$, $\mathcal{M}_{\rm int}$, the past trajectory encoders  $\mathcal{E}_\mathrm{social}(\cdot)$, $\mathcal{E}_\mathrm{int}(\cdot)$ and the decoder $\mathcal{D}(\cdot)$. Train the memory addresser with the loss $\mathcal{L}_\mathrm{Addr}$.

4. Freeze the MemoNet and train the trajectory fulfillment encoder 
$\mathcal{E}_\mathrm{full}(\cdot)$ and decoder $\mathcal{D}_\mathrm{full}(\cdot)$ with the loss $\mathcal{L}_\mathrm{traj}$.

\begin{table*}[!t]
\footnotesize
\centering
\setlength{\tabcolsep}{1mm}{\caption{\small minADE$_{20}$ / minFDE$_{20}$ (pixels) of trajectory prediction (SDD dataset). Lower is better. The bold/underlined font represent the best/second best result. Our method achieves a \textbf{20.3\%} FDE improvement compared to PECNet.}
\vspace{-10pt}
\fontsize{8.5}{10.5}\selectfont
\begin{tabular}{l|ccccccccc|c}
\hline
\hline
    Time 
    & \makecell[c]{Social\\-GAN\cite{gupta2018social}}
    &\makecell[c]{Social-\\STGCNN\cite{mohamed2020social}}
      & \makecell[c]{Trajectron++\\\cite{salzmann2020trajectron++}}
      &\makecell[c]{SOPHIE\\\cite{sadeghian2019sophie}}
    & \makecell[c]{NMMP\cite{hu2020collaborative}}
    & \makecell[c]{EvolveGraph\\\cite{li2020evolvegraph}}&CF-VAE \cite{bhattacharyya2019conditional}
    &\makecell[c]{MANTRA\\ \cite{marchetti2020mantra}}  
    &\makecell[c]{PECNet\\\cite{mangalam2020not}}
    &\textbf{Ours}
    \\
\hline
\rule{0pt}{9pt}
     4.8s& 27.23/41.44 & 20.60/33.10  &19.30/32.70&16.27/29.38&14.67/26.72&13.90/22.90&12.60/22.30&\underline{8.96}/17.76&9.96/\underline{15.88} &\textbf{8.56}/\textbf{12.66}\\
\hline
\end{tabular}
\label{table:sdd}}
\vspace{-3.5mm}
\end{table*}

\begin{table*}[!t]
\centering
\setlength{\tabcolsep}{1mm}\caption{minADE$_{20}$ / minFDE$_{20}$ (meters) of trajectory prediction (ETH-UCY dataset). Lower is better. The bold/underlined font represent the best/second best result. Our method achieves a \textbf{10.2\%} FDE improvement compared to Agentformer.}
\vspace{-10pt}
\fontsize{8.5}{10.5}\selectfont
\resizebox{\textwidth}{!}{
\begin{tabular}{l|ccccccccc|c}
  \hline
  \hline
    Subset & \makecell[c]{Social-\\GAN\cite{gupta2018social}}&\makecell[c]{STGAT \cite{huang2019stgat}}&NMMP\cite{hu2020collaborative}
    &\makecell[c]{MANTRA\\ \cite{marchetti2020mantra}}
    &\makecell[c]{Transformer\\-TF \cite{giuliari2021transformer}}
    &STAR\cite{yu2020spatio}&\makecell[c]{PECNet\\\cite{mangalam2020not}} &\makecell[c]{Trajectron++\\\cite{salzmann2020trajectron++}} &\makecell[c]{Agentformer\\\cite{yuan2021agentformer}}
    &\makecell[c]{\textbf{Ours}}\\
\hline
     ETH& 0.87/1.62&0.65/1.12&0.61/1.08&0.48/0.88& 0.61/1.12&\textbf{0.36}/\underline{0.65}&0.54/0.87 &\underline{0.39}/0.83&0.45/0.75 &0.40/\textbf{0.61}\\
     HOTEL&0.67/1.37&0.35/0.66&0.33/0.63&0.17/0.33&0.18/0.30&0.17/0.36&0.18/0.24&\underline{0.12}/\underline{0.21}&0.14/0.22&\textbf{0.11}/\textbf{0.17}\\
     UNIV&0.76/1.52&0.52/1.10&0.52/1.11&0.37/0.81&0.35/0.65&0.31/0.62&0.35/0.60&\textbf{0.20}/\underline{0.44}&0.25/0.45&\underline{0.24}/\textbf{0.43}  \\
     ZARA1&0.35/0.68&0.34/0.69&0.32/0.66&0.27/0.58&0.22/0.38&0.29/0.52&0.22/0.39 &\textbf{0.15}/0.33&\underline{0.18}/\textbf{0.30}&\underline{0.18}/\underline{0.32}\\
     ZARA2&0.42/0.84&0.29/0.60&0.43/0.85&0.30/0.67&0.17/0.32&0.22/0.46&0.17/0.30&\textbf{0.11}/0.25 &\underline{0.14}/\textbf{0.24}&\underline{0.14}/\textbf{0.24}\\
     AVG&0.61/1.21&0.43/0.83&0.41/0.82&0.32/0.65&0.31/0.55&0.26/0.53&0.29/0.48&\textbf{0.19}/0.41 &0.23/\underline{0.39}&\underline{0.21}/\textbf{0.35}\\
  \hline
\end{tabular}
\label{table:eth}}
\vspace{-3.5mm}
\end{table*}

\begin{table*}[!t]
\footnotesize
\centering
\setlength{\tabcolsep}{1mm}{\caption{\small minADE$_{20}$ / minFDE$_{20}$ (meters) of trajectory prediction (NBA dataset). Lower is better. The bold/underlined font represent the best/second best result. Our method achieves a \textbf{28.3\%} FDE improvement compared to NMMP.}
\vspace{-10pt}
\fontsize{8.5}{10.5}\selectfont
\begin{tabular}{l|p{1.6cm}<{\centering} p{1.65cm}<{\centering} p{1.85cm}<{\centering} p{1.75cm}<{\centering} p{1.25cm}<{\centering} p{1.75cm}<{\centering} p{1.75cm}<{\centering} p{1.75cm}<{\centering} | p{1.3cm}<{\centering}}
\hline
\hline
    Time 
    & \makecell[c]{Social-\\LSTM\cite{alahi2016social}}
    & \makecell[c]{Social-\\GAN\cite{gupta2018social}}
    &\makecell[c]{Social-\\STGCNN\cite{mohamed2020social}}
    &\makecell[c]{STGAT \cite{huang2019stgat}}
    & NRI\cite{kipf2018neural}
    & \makecell[c]{STAR\cite{yu2020spatio}}
    &\makecell[c]{PECNet\cite{mangalam2020not}}
    & \makecell[c]{NMMP\cite{hu2020collaborative}}
    &\textbf{Ours}
    \\
\hline
\rule{0pt}{9pt}
     4.0s& 1.79/3.16 & 1.62/2.51  &1.59/2.37&1.41/2.22&2.06/3.74&1.26/2.04&1.83/3.41&\underline{1.33}/\underline{2.05} &\textbf{1.25}/\textbf{1.47}\\
\hline
\end{tabular}
\label{table:nba}}
\vspace{-4mm}
\end{table*}

\section{Experiments}
\subsection{Datasets}
\vspace{-1mm}
\textbf{Stanford Drone Dataset (SDD)}: SDD is a large-scale dataset collected from campus in bird's eye view. Following~\cite{gupta2018social,mangalam2020not}, we use the standard train-test split and predict the future 4.8s (12 frames) using past 3.2s (8 frames).

\textbf{ETH-UCY}: The ETH-UCY dataset contains 5 subsets, including ETH, HOTEL, UNIV, ZARA1 and ZARA2 containing various scenes captured at 2.5Hz. We use the same segment length of 8s as SDD following previous works~\cite{hu2020collaborative,mangalam2020not} and use the leave-one-out approach with 4 sets for training and the remaining set for testing.

\textbf{NBA}: The NBA trajectory dataset is collected by NBA using the SportVU tracking system, which reports the trajectories of the ten players and the ball in real basketball games. We randomly sample 50k samples for training and testing.

\vspace{-1mm}

\subsection{Implementation details}
\vspace{-1mm}
For MemoNet, the feature dimensions of the past memory bank and the intention memory bank are 128 and 64, respectively. On SDD, we filter the initial memory banks with $\theta_{\rm past}=1$, $\theta_{\rm int} = 1$ and the coarse intention anchor number $L$ is 120. On ETH-UCY, we filter the initial memory banks with $\theta_{\rm past}=0.02$, $\theta_{\rm int} = 0.02$ and the coarse intention anchor number $L$ is 320. The coefficients $\alpha$ and $\beta$ in loss functions are set to 1. We train the entire framework with SGD optimizer~\cite{Bottou10}. We use an initial learning rate of $10^{-3}$ to train the feature learning framework, $10^{-4}$ to train the memory addresser, and $10^{-3}$ to train the trajectory fulfillment. All these modules are finetuned with a learning rate of $10^{-6}$. See more details in the supplementary material.
\vspace{-1mm}

\subsection{Quantitative results}
\vspace{-1mm}
Two used evaluation metrics are the minimum average displacement error ($\mathrm{minADE}_K$), which is the minimum among $K$ time-averaged distances of predicted trajectories compared to the ground-truths, and the minimum final displacement error ($\mathrm{minFDE}_K$), which is the minimum distance among $K$ predicted endpoints to the ground-truth endpoints.

On SDD dataset, we compare our method with current $9$ state-of-the-art prediction methods; see Table \ref{table:sdd}.
We see that i) our MemoNet significantly outperforms all baselines in intention prediction measured by FDE. Our method reduces FDE from 15.88 to 12.66 compared to the current state-of-the-art method, PECNet, achieving \textbf{20.3\%} improvement; ii) with a more precise intention prediction, our method predicts the whole trajectory more accurately. Our method outperforms PECNet by \textbf{14.1\%} in ADE. 

On ETH-UCY dataset, we compare our method with $9$ prediction methods; see Table \ref{table:eth}. We see that i) MemoNet outperforms competitive methods in predicting intention measured by FDE. Specifically, our method reduces the average FDE from 0.39 to 0.35 compared to the previous state-of-the-art method, AgentFormer, achieving \textbf{10.2\%} improvement; and ii) our method achieves the best or close to the best performance in ADE over all the five subsets.

On NBA dataset, we compare our proposed method with $8$ prediction methods; see Table \ref{table:nba}. We see that MemoNet reduces FDE from 2.05 to 1.47 compare to the current state-of-the-art method, NMMP, achieving \textbf{28.3\%} improvement.


\begin{figure}[t] 
\centering
\includegraphics[width=0.48\textwidth]{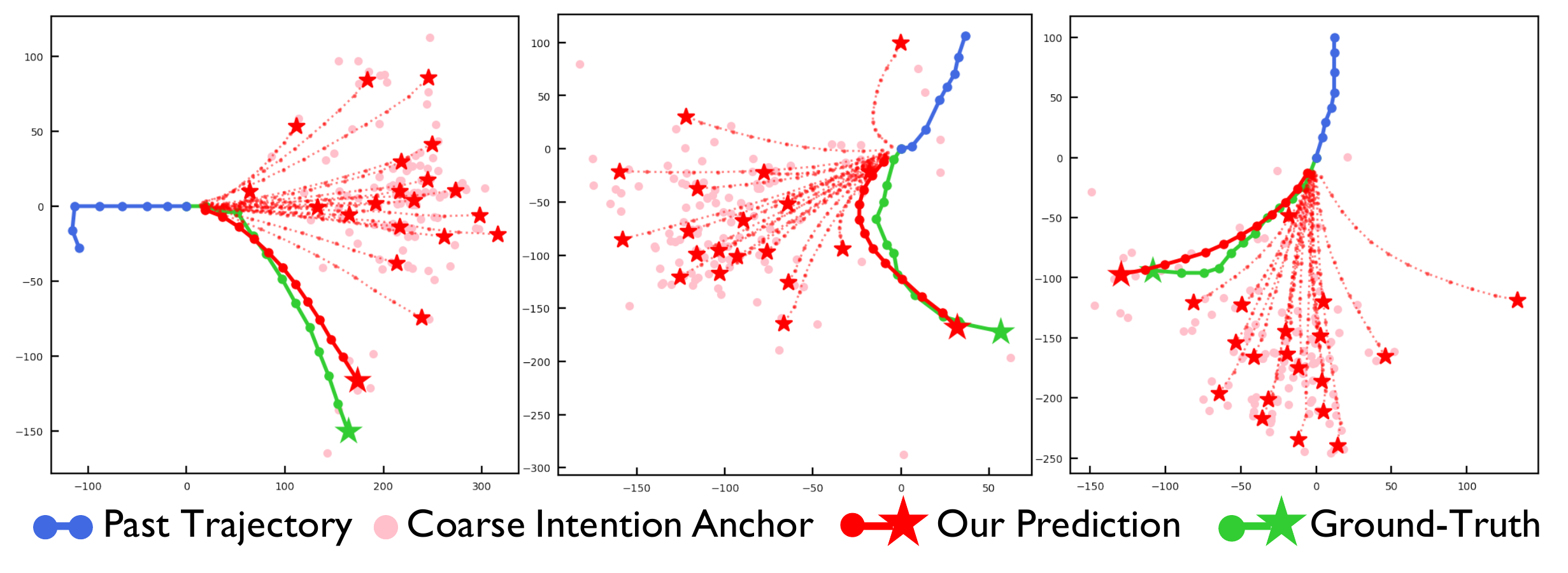}
\vspace{-7mm}
\caption{\small Diverse intention prediction by MemoNet on SDD, where 20 final intentions are clustered from 120 coarse intention anchors. MemoNet can provide diverse and accurate intention predictions.}
\label{fig:sdd_clustering}
\vspace{-0pt}
\end{figure}

\subsection{Qualitative results}
\vspace{-1mm}
\textbf{Visualization of diverse intention.} 
Fig.\ref{fig:sdd_clustering} illustrates the diverse intention prediction with MemoNet, where the pink dots are coarse intention anchors. We see that with the help of intention clustering, MemoNet can provide diverse and accurate intention predictions.

\begin{figure}[t] 
\centering
\includegraphics[width=0.49\textwidth]{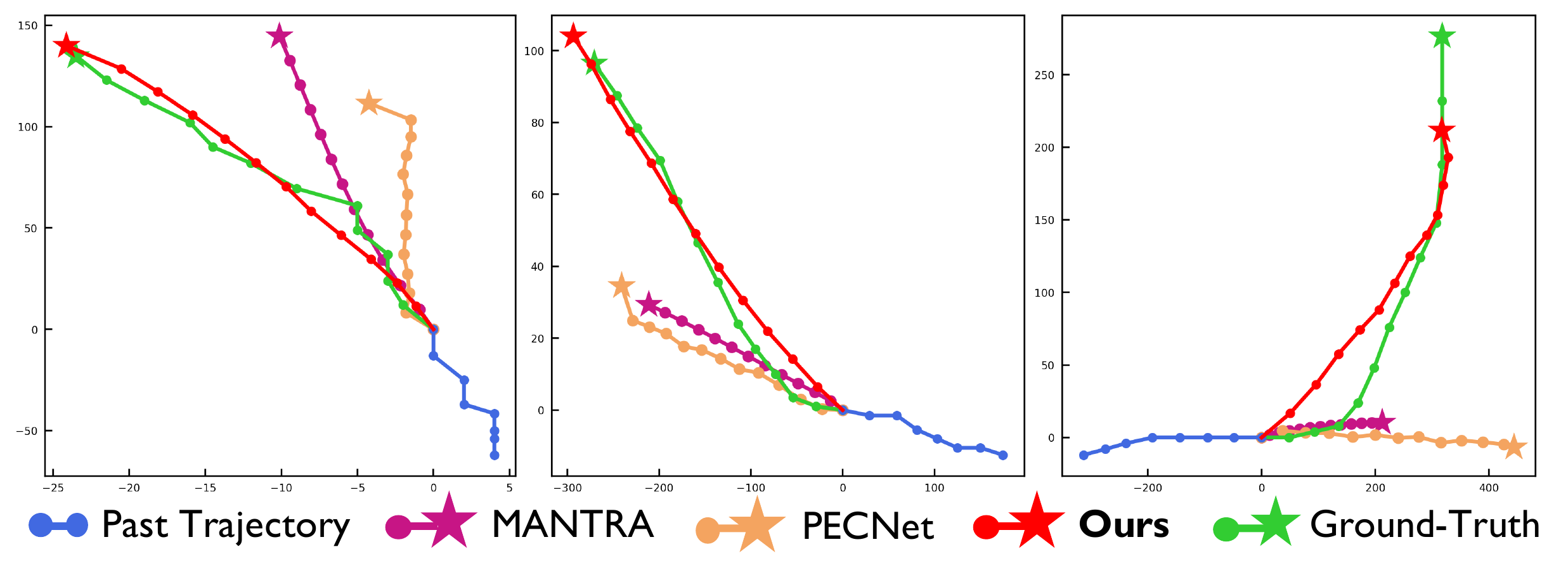}
\vspace{-7.5mm}
\caption{\small We compare the best-of-20 predicted trajectories produced by our method and two previous methods on SDD. Our method achieves a more precise trajectory prediction. }
\label{fig:comparison}
\vspace{-10pt}
\end{figure}

\textbf{Visualization of predicted trajectory.} Fig.\ref{fig:comparison} compares the best-of-20 predicted trajectories produced by our MemoNet and previous state-of-the-art method PECNet and MANTRA. We see that our predictions (red) are closer to the ground-truth (green) than other two methods. Especially, for challenging direction-turning cases (third column), previous methods fail to capture the right direction; while our MemoNet still provides precise prediction.

\begin{figure}[t] 
\centering
\includegraphics[width=0.49\textwidth]{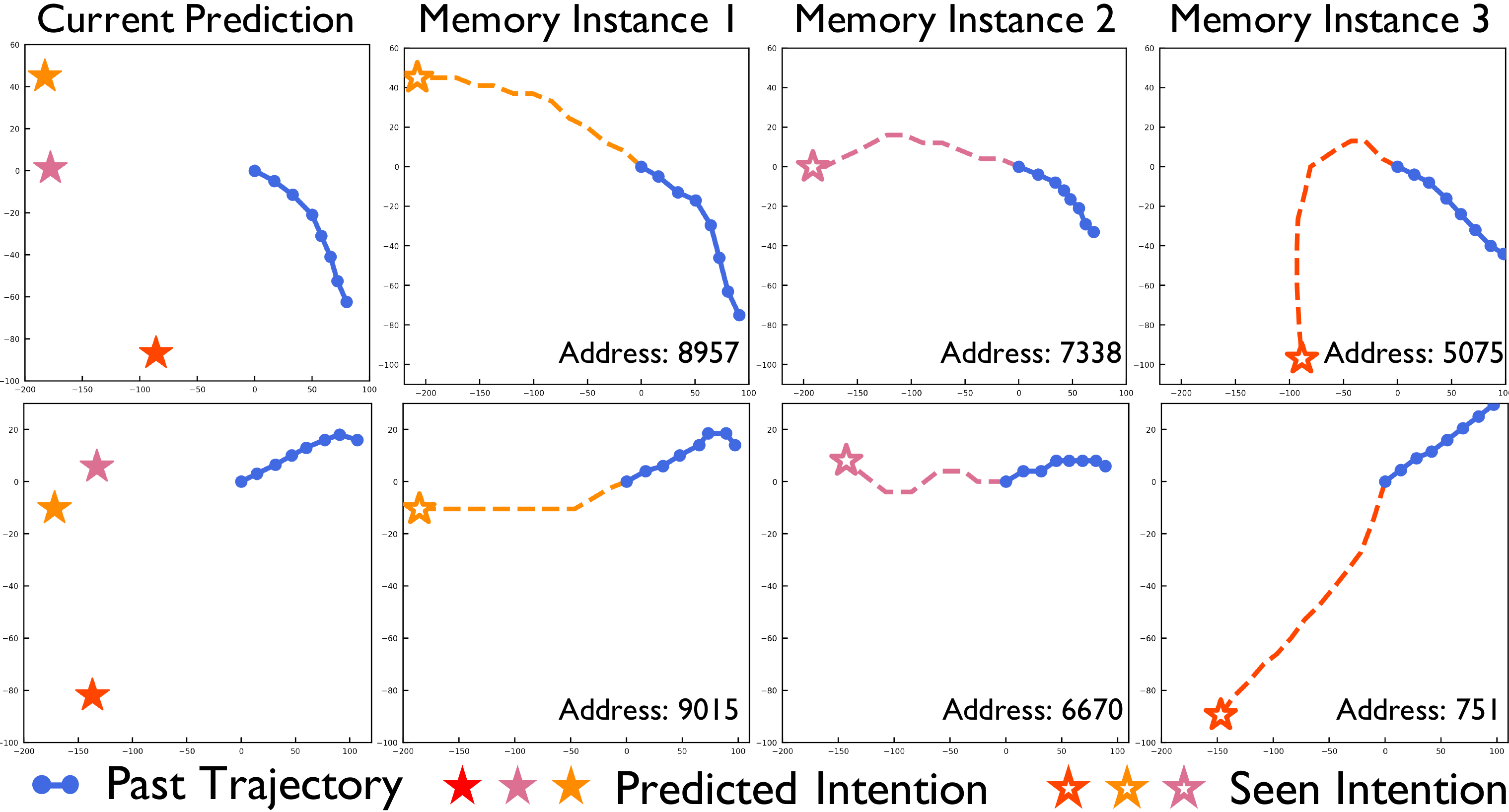}
\vspace{-7.5mm}
\caption{\small Prediction cases with corresponding past-future trajectories traced by memory addresser. Our model promotes a more explicit link between the current situation and seen instances.}
\label{fig:intention}
\vspace{-10pt}
\end{figure}

\textbf{Visualization of explicit link.}
Fig.\ref{fig:intention} shows prediction cases with their seen past-future trajectory instances traced by the addressed memory instances. We see that seen similar scenarios provide instance-level experience to obtain multimodal future intentions and reflects that our model can trace back to specific memorized samples during the prediction.

\setlength{\tabcolsep}{3pt}
\begin{table}[t]
\footnotesize
\centering
\renewcommand{\arraystretch}{1.2}
\caption{\small Ablation study of each component in MemoNet on the SDD and ETH dataset. $\circ/\checkmark$ represent using cosine distance/learnable addresser. Each component is beneficial.}
\vspace{-3mm}
\begin{tabular}{cccc|c|c}
\hline
\hline
\rule{0pt}{10pt} 

\makecell[c]{ Memory\\Bank} & \makecell[c]{Memory\\Filtering} & \makecell[c]{Memory\\Addresser} & \makecell[c]{Intention\\Clustering} & SDD & ETH \\
\hline
   & &   &    & 14.16/27.76          & 0.78/1.44\\
 \checkmark &   &  $\circ$&    & 9.64/15.25           & 0.55/0.94\\
 \checkmark &  \checkmark  &  $\circ$&    & 9.59/15.08           & 0.55/0.93\\
 \checkmark &  \checkmark  &  \checkmark    &  & 9.50/14.78           & 0.53/0.89\\
 \checkmark &  \checkmark  &  \checkmark  &  \checkmark  & \textbf{8.56}/\textbf{12.66}           & \textbf{0.40}/\textbf{0.61}\\
\hline
\hline
\end{tabular}
\label{table:ablation_on_sdd}
\vspace{0mm}
\end{table}

\subsection{Ablation studies}
\vspace{-1mm}
\textbf{Effect of components in MemoNet.} 
We explore the effect of each of four proposed key components in MemoNet, including memory bank, memory filtering, memory addresser and intention clustering. Table~\ref{table:ablation_on_sdd} presents the results. We see that i) the proposed memory bank can significantly improve the prediction performance; and ii) the memory filtering, learnable addresser and intention clustering all contribute to promoting accurate prediction.

\begin{figure}[t] 
\centering
\includegraphics[width=0.47\textwidth]{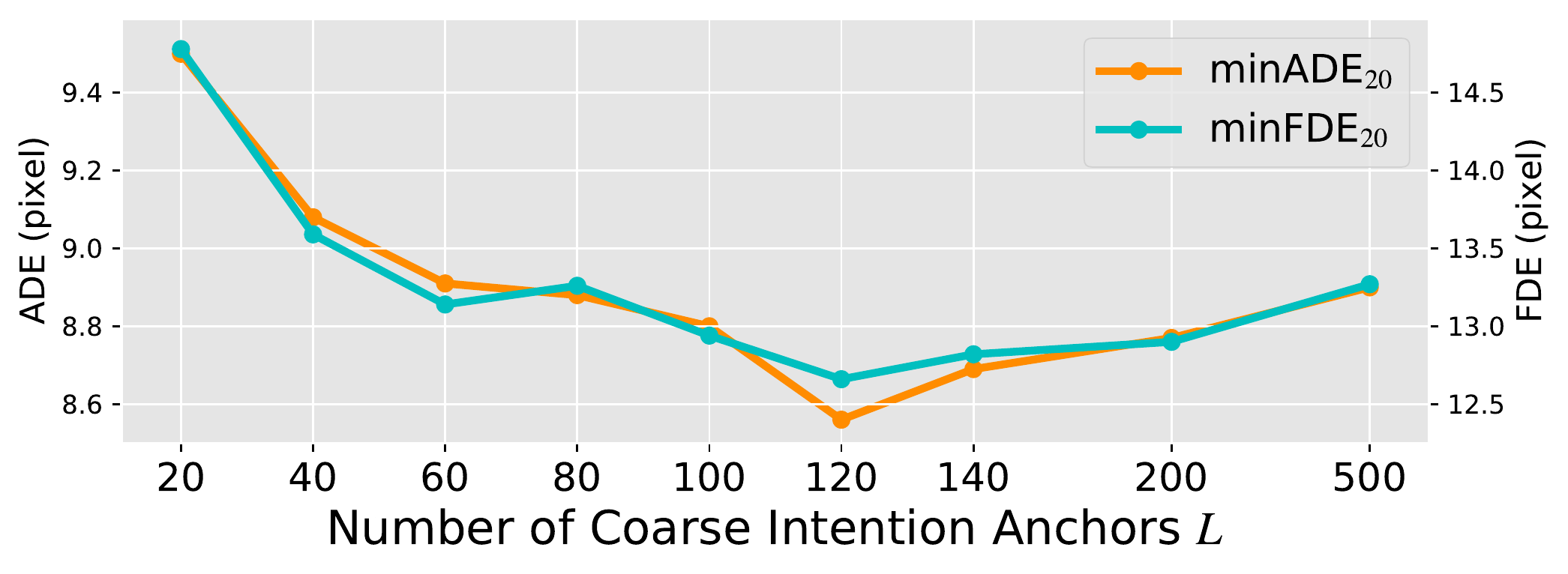}
\vspace{-3mm}
\caption{\small ADE/FDE as a function of the number of coarse intention anchors $L$ on SDD. $L=120$ provides the best performance.}
\label{fig:sdd_number_of_destination}
\vspace{-0pt}
\end{figure}

\textbf{Effect of the number of coarse intention anchors.}
Fig.\ref{fig:sdd_number_of_destination} illustrates the influence of coarse intention anchor numbers $L$. We find that either too small or too large $L$ causes performance degeneration as i) when $L$ is small, the model tends to miss intention modality, causing insufficient diversity and worse prediction performance; and ii) when $L$ is too large, the  prediction involves too many irrelevant instances, also resulting in worse prediction performance.

\setlength{\tabcolsep}{3pt}
\begin{table}[t]
\footnotesize
\centering
\renewcommand{\arraystretch}{1.1}
\caption{\small Ablation study of thresholds $\theta_{\mathrm{past}}/\theta_{\mathrm{int}}$ in memory filtering on SDD. $\theta_{\mathrm{past}}=\theta_{\mathrm{int}}=1$ achieves the best performance.}
\vspace{-3mm}
\begin{tabular}{c|c|r|r}
\hline
\hline
\makecell[c]{$\theta_{\mathrm{past}}/\theta_{\mathrm{int}}$}& {minADE$_{20}$}/{minFDE$_{20}$}& \makecell[c]{Memory Size} &\;Storage\\
\hline
0       & 8.65/12.84        & 17970 (100.0\%)          &   13.8MB \\
0.5     & 8.59/12.70        & 15442 (85.9\%)           & 11.9MB \\
1       & \textbf{8.56/12.66}  & 14652 (81.5\%)        & 11.2MB \\
5       & 9.22/14.29        & 10698 (59.5\%)        & 8.2MB \\
10      & 9.64/15.57        & 6635  (36.9\%)        & 5.1MB \\
20      & 10.41/17.32       & 2692  (15.0\%)        & 2.1MB \\
50      & 13.77/25.86       & 604  (3.4\%)      & 465KB \\
\hline
\hline
\end{tabular}
\label{table:ablation_threshold_on_sdd}
\vspace{-4mm}
\end{table}

\textbf{Effect of thresholds in memories filtering.} Table \ref{table:ablation_threshold_on_sdd} reports the prediction errors with various thresholds $\theta_{\mathrm{past}}/\theta_{\mathrm{int}}$ in memory filtering. We see that i) an appropriate $\theta_{\mathrm{past}}/\theta_{\mathrm{int}}$ leads to a remarkable performance and lightweight storage; ii) when $\theta_{\mathrm{past}}/\theta_{\mathrm{int}}$ are too small, the model tends to preserve redundant information and decrease the intention diversity, wasting the storage and affecting the performance; and iii) when $\theta_{\mathrm{past}}/\theta_{\mathrm{int}}$ are too large, a large amount of useful information are filtered out, which makes it harder to find relevant instances, deteriorating the performance.

\textbf{Real-time inference speed.} We run the whole inference model for 10 times on SDD dataset using one RTX-3090 GPU. The average prediction time is 18.03ms per sample, with a real-time predictions FPS=55.5, much faster than the common sampling rate of data collection. 

\vspace{-1mm}
\section{Conclusion}
\vspace{-1mm}
This paper proposes MemoNet, an instance-based approach that is designed based on the retrospective memory mechanism, where the seen instances are stored into a memory bank pair during training and could be used for relevant movement pattern matching during inference. The proposed MemoNet includes four key designs: a joint-reconstruction-based feature-learning architecture, a memory filtering algorithm, a learnable addresser, and an intention clustering method. Experiments show that our method significantly improves the state-of-the-art performance on trajectory prediction datasets and has the ability to trace back to specific instances during prediction, promoting more interpretability.

\textbf{Limitation and future work.} In this paper, we focus on memorizing past-intention pairs. However, it is a challenge to predict some special actions only using past trajectories, such as a sharp turn. A future work is to utilize the map information to generate environment conditioned prediction.

\section*{Acknowledgements}
\vspace{-2mm}
This research is partially supported by the National Key R\&D Program of China under Grant 2021ZD0112801, National Natural Science Foundation of China under Grant 62171276, the Science and Technology Commission of Shanghai Municipal under Grant 21511100900 and CCF-DiDi GAIA Research Collaboration Plan 202112. 



{\small
\bibliographystyle{ieee_fullname}
\bibliography{main}
}

\end{document}